\documentclass[11pt]{article}
\usepackage[preprint]{acl}

\usepackage{times}
\usepackage{latexsym}

\usepackage[T1]{fontenc}

\usepackage[utf8]{inputenc}

\usepackage{microtype}

\usepackage{inconsolata}

\usepackage{graphicx}

\usepackage{multirow}
\usepackage[normalem]{ulem}

\usepackage[table]{xcolor}
\usepackage{booktabs}
\usepackage{array}
\usepackage{pifont}

\newcommand{\cmark}{\ding{51}}
\newcommand{\xmark}{\ding{55}}

\definecolor{ForestGreen}{RGB}{34,139,34}

\newcommand{\bad}[1]{\textcolor{red}{#1}}

\title{A Diagnostic Study of Faithfulness in Diffusion Language Models for Table-to-Text Generation}

\author{
  \textbf{Zaynab Merimi\textsuperscript{1,*}},
  \textbf{Noussayba El Marrakchi\textsuperscript{1,*}},
  \textbf{Arthur Satouf\textsuperscript{2,3}}
\\[6pt]
  \textsuperscript{1}CY Tech, CY Cergy Paris Université, France
\\
  \textsuperscript{2}MILA -- Quebec AI Institute, Canada
\\
  \textsuperscript{3}Université Paris-Saclay \& CentraleSupélec, France
\\[6pt]
  \small{
    \textsuperscript{*}Equal contribution
  }
  \\
    \small{
  \texttt{zaynabmerimi15(at)gmail.com},
  \texttt{noussayba.el.marrakchi(at)gmail.com},
  \texttt{arthur.satouf(at)gmail.com}
  }
}

\begin{document}
\maketitle
\raggedbottom
\begin{abstract}

Diffusion language models generate by iteratively denoising a masked sequence under bidirectional attention, without the strict left-to-right commitment of autoregressive decoding. Whether this difference affects how reliably they remain grounded in structured inputs is largely unexplored. We study this question in table-to-text generation using ToTTo, where highlighted source cells make factual grounding directly testable. We compare the masked diffusion models MDLM and LLaDA with GPT-2 and GPT-Neo under an identical retrieval-based few-shot protocol, together with a QLoRA-adapted LLaDA, and we score reference similarity and source grounding as separate axes. The two pull apart. Within LLaDA, additional demonstrations raise BLEU from 11.3 to 22.0 while NLI entailment against the source table falls from 0.78 to 0.52; QLoRA reaches the best BLEU of the study, 24.15, and grounds no better for it. Manual annotation shows the failure modes differ in kind: the autoregressive baselines tend to wander off the table, while LLaDA-based models more often stay on topic and swap out individual entities. Reference-based metrics alone are insufficient to assess faithfulness in table-to-text generation.

\end{abstract}

\section{Introduction}

Tables are among the most common structured elements found in real-world documents, from financial reports to scientific papers, making faithful table-to-text generation a prerequisite for reliable document-grounded reasoning. Natural language generation has long been dominated by autoregressive language models, which generate text sequentially from left to right~\cite{Radford2019GPT2}. Diffusion-based language models have recently emerged as a fundamentally different alternative, generating text through iterative denoising under bidirectional attention~\cite{Kocabay2026Diffutron}. MDLM~\cite{Sahoo2024MDLM} first showed that masked diffusion models could approach autoregressive perplexity on language-modeling benchmarks; LLaDA~\cite{Nie2025LLaDA} extended this line of work well beyond perplexity. Nevertheless, one question remains unexplored: can diffusion language models be trusted to generate text that is faithful to a structured input source?

To investigate this question, we need a setting where faithfulness violations are unambiguous and objectively verifiable. Table-to-text generation provides exactly this testbed. Early table-to-text benchmarks such as WikiBio~\cite{Lebret2016WikiBio} suffered from reference divergence from the source table~\cite{Dhingra2019Parent}. ToTTo~\cite{Parikh2020ToTTo} addressed this by having models describe only a highlighted subset of table cells. Prior work on ToTTo, however, has relied on fine-tuned models such as T5~\cite{Kale2020T5Totto} and BART~\cite{Lewis2020BART}, evaluated with lexical metrics. Krylov and Somov~\cite{Krylov2024GENIE} adapt the continuous encoder--decoder diffusion model GENIE and show it matches fine-tuned T5 baselines on BLEU, PARENT and BLEURT while offering a better quality--diversity trade-off. Their study, however, treats table grounding as one quality score among several. We ask a different question. We take faithfulness as the object of study rather than a metric.

Hallucination is commonly split into intrinsic errors (contradicting the source) and extrinsic errors (introducing unsupported content)~\cite{Ji2023Survey}. Lexical metrics like BLEU/ROUGE reward reference similarity regardless of grounding~\cite{Dhingra2019Parent}, motivating our use of NLI-based entailment~\cite{Laban2022Summac} as a source-grounded faithfulness score.

\cite{Piskorz2025Masks} raise concerns that the training objective underlying masked diffusion alters the inductive biases familiar from autoregressive models in ways that may limit a model's ability to attend to and faithfully reproduce specific input content. 

We make three contributions: (i) we evaluate masked discrete diffusion models (MDLM, LLaDA) rather than continuous-space diffusion, in the frozen few-shot regime rather than after task-specific fine-tuning; (ii) we score reference similarity and source grounding on separate axes and show they diverge systematically -- within a single model, adding in-context demonstrations improves BLEU while reducing entailment against the source table, and QLoRA improves aggregate lexical scores without yielding statistically significant gains in source faithfulness after correction for multiple comparisons; and (iii) we characterize how the paradigms fail, not just how often, showing a shift from coarse topic drift in the autoregressive baselines to localized entity substitution in LLaDA-based generations.

\section{Experimental Setup and Results}

\subsection{Dataset}

We conduct experiments on the ToTTo benchmark introduced by ~\cite{Parikh2020ToTTo} and only the highlighted cells were retained. We transformed each structured table into a textual representation prior to inference. Each processed instance consists of three fields: (i) a table identifier, (ii) the serialized table representation, and (iii) the reference description from the dataset.

We constructed a training subset containing 5,000 examples and a validation subset containing 500 examples, extracted from the official ToTTo training and development splits, respectively.

\subsection{Language Models}

We explore whether masked diffusion language models can produce faithful descriptions of structured data. We first tested MDLM ~\cite{Sahoo2024MDLM} before focusing on LLaDA ~\cite{Nie2025LLaDA}. We evaluate the two models in a few-shot setting with $k \in \{0,1,3,5,10\}$ in-context examples ($k=0$: zero-shot), each constructed by prepending $k$ randomly sampled ``Table/Description'' pairs to the target table. We then fine-tune LLaDA on the training set.

We study then the text generation from representative autoregressive baselines. Table~\ref{tab:model_summary} summarizes the characteristics of the evaluated models. All models are evaluated under the same prompting protocol and in-context learning settings.
\begin{table}[h]
\centering
\caption{Summary of the evaluated language models.}
\label{tab:model_summary}
\small
\begin{tabular}{lccc}
\toprule
Model & Architecture & Parameters \\
\midrule
GPT-2 & Autoregressive & 117M & \xmark \\
GPT-Neo & Autoregressive & 1.3B & \xmark \\
MDLM & Diffusion & 0.6B & \xmark \\
LLaDA & Diffusion & 8B & \xmark \\
LLaDA + QLoRA & Diffusion & 8B & \cmark \\
\bottomrule
\end{tabular}
\end{table}

We evaluate outputs with five complementary metrics: BLEU-4 and
ROUGE-L for lexical overlap, BERTScore for semantic similarity,
Self-BLEU for output diversity, and source-grounded NLI entailment
for faithfulness. We complement these automatic metrics with
qualitative evaluation.

\subsection{Quantitative Evaluation}
\begin{table*}[t]
\centering
\caption{Best performance achieved by each model across all in-context learning settings, selected according to BLEU. Best results are highlighted in bold and second-best results are underlined. Lower Self-BLEU indicates greater generation diversity.}
\label{tab:best_results}
\small
\begin{tabular}{lcccccc}
\toprule
\textbf{Model} & \textbf{Best Shot} & \textbf{BLEU $\uparrow$} &
\textbf{R-L $\uparrow$} & \textbf{BERTScore $\uparrow$} &
\textbf{NLI Ent. $\uparrow$} & \textbf{Self-BLEU $\downarrow$} \\
\midrule

GPT-2
& 1
& 4.64
& 0.212
& 0.837
& 0.239
& \underline{0.085} \\

GPT-Neo
& 1
& 5.74
& 0.201
& 0.834
& 0.317
& 0.092 \\

MDLM
& 10
& 7.93
& 0.267
& 0.875
& 0.191
& 0.088 \\

LLaDA
& 5
& \underline{21.96}
& \underline{0.468}
& \textbf{0.917}
& \textbf{0.516}
& 0.088 \\

LLaDA + QLoRA
& 1
& \textbf{24.15}
& \textbf{0.474}
& \underline{0.914}
& \underline{0.355}
& \textbf{0.059} \\

\bottomrule
\end{tabular}
\end{table*}
Table~\ref{tab:all_results} reports the performance of all models
across the five in-context learning (ICL) settings, while
Table~\ref{tab:best_results} summarizes the best configuration of each
model according to BLEU. Several trends emerge.

First, LLaDA-based models substantially outperform the autoregressive
baselines and MDLM on reference-based metrics. The best BLEU score is
obtained by LLaDA+QLoRA at 1-shot (24.15), followed by pretrained
LLaDA at 5-shot (21.96), compared with 7.93 for MDLM, 5.74 for
GPT-Neo, and 4.64 for GPT-2. LLaDA+QLoRA also achieves the highest
ROUGE-L score among the BLEU-selected configurations (0.474), while
LLaDA obtains the highest BERTScore (0.917). These results indicate a
substantial advantage of the LLaDA-based models in matching the
lexical and semantic content of the reference descriptions.

The effect of in-context learning varies markedly across models.
GPT-2 and GPT-Neo reach their highest BLEU scores with a single
demonstration and do not benefit consistently from additional
examples. MDLM exhibits a different pattern, improving from 4.73 BLEU
in the zero-shot setting to 7.93 at 10-shot. LLaDA shows the largest
initial gain: BLEU increases from 11.25 at zero-shot to 21.54 with
one demonstration, after which performance remains relatively stable.
QLoRA adaptation further increases the best BLEU to 24.15, although
additional demonstrations do not provide monotonic improvements.
Overall, these results suggest that the benefit of ICL is strongly
model-dependent and that, for LLaDA-based models, most of the
reference-based gain is obtained with only a small number of
demonstrations.

Self-BLEU reveals a complementary pattern. The lowest value among the
BLEU-selected configurations is obtained by LLaDA+QLoRA (0.059),
indicating that its improved reference overlap is not accompanied by
greater repetition across outputs. However, diversity alone does not
establish factual correctness and must therefore be considered
alongside source-grounded evaluation.

NLI-based evaluation substantially changes the interpretation of the
reference-based results. When the serialized source table is used as
the premise and the generated description as the hypothesis, higher
BLEU and ROUGE do not systematically correspond to higher
entailment. Most notably, the highest entailment observed across all
settings is obtained by pretrained LLaDA in the zero-shot condition
(0.781), followed by zero-shot LLaDA+QLoRA (0.726). Increasing the
number of demonstrations generally improves reference similarity for
LLaDA while reducing entailment relative to its zero-shot result.
Likewise, the BLEU-optimal LLaDA+QLoRA configuration reaches 24.15
BLEU but only 0.355 entailment, compared with 0.516 entailment for
the BLEU-optimal pretrained LLaDA configuration. This divergence
shows that improved agreement with the human reference should not be
interpreted as equivalent to improved grounding in the source table.

\paragraph{Statistical significance.}
To determine whether the gains obtained with QLoRA reflect systematic
differences rather than variation in aggregate scores, we performed
paired two-sided Wilcoxon signed-rank tests between pretrained LLaDA
and LLaDA+QLoRA within each ICL setting. Predictions were paired by
evaluation example. We tested sentence-level BLEU, ROUGE-L,
BERTScore, NLI entailment, and NLI contradiction, applying Holm
correction for multiple comparisons within each setting.

After correction, QLoRA does not yield a statistically significant
improvement over pretrained LLaDA on sentence-level BLEU, ROUGE-L,
or BERTScore at any evaluated ICL setting
($p_{\mathrm{Holm}} > 0.05$). Differences instead emerge for
faithfulness-related metrics in selected settings. Most notably, at
1-shot, pretrained LLaDA achieves substantially higher NLI
entailment (0.630 vs.\ 0.355,
$p_{\mathrm{Holm}} < 10^{-5}$) and lower NLI contradiction
(0.013 vs.\ 0.068, $p_{\mathrm{Holm}} < 0.001$) than
LLaDA+QLoRA. Thus, although QLoRA yields the highest aggregate BLEU
and ROUGE-L scores, these gains are not statistically robust at the
example level under the tested reference-based metrics and do not
translate into improved source faithfulness.

\subsection{Qualitative Analysis and Human Evaluation}

We complement automatic evaluation with a human annotation study on 30 examples 
per model at $k=3$, categorizing outputs into four error types: \textit{Faithful}, 
\textit{Entity Substitution}, \textit{Topic Drift or Prompt Leakage}, and
\textit{Other Unsupported Content}. 
Results are reported in Table~\ref{tab:human_eval}.

 GPT-2 is dominated 
by topic drift (40\%), indicating that few-shot prompting alone fails to anchor 
its outputs to the source table. GPT-Neo shows a lower topic drift rate (30\%),
although its generations still exhibit entity
substitutions and unsupported content.

LLaDA achieves the highest faithful rate (47\%) and the lowest topic drift 
(10\%), indicating stronger conditioning on the structured
input among the evaluated models. However, fine-tuning with QLoRA 
unexpectedly degrades faithfulness to 26.67\% while increasing entity 
substitutions to 40\% — suggesting that task adaptation improves format 
adherence but introduces localized factual errors, suggesting that task adaptation improves format adherence while
remaining susceptible to localized factual errors.

These results reveal a qualitative shift in error type across paradigms: 
autoregressive models exhibit coarse-grained failures such as topic drift, 
while diffusion models produce more subtle, localized errors such as entity 
substitution.

\section{Discussion and Limitations}

Our experiments provide, to our knowledge, the first empirical
comparison between this set of autoregressive and diffusion language
models on a structured table-to-text benchmark from the perspective
of faithfulness. LLaDA-based models achieve substantially higher
BLEU, ROUGE, and BERTScore than the considered autoregressive
baselines and MDLM. QLoRA further improves the best aggregate BLEU
and ROUGE-L scores. However, paired statistical testing shows that
these improvements over pretrained LLaDA are not significant at the
example level for sentence-level BLEU, ROUGE-L, or BERTScore after
Holm correction.

More importantly, improved reference similarity does not systematically
imply improved source grounding. Source-grounded NLI evaluation
reveals that pretrained LLaDA can achieve substantially higher
entailment than its QLoRA-adapted counterpart despite lower aggregate
BLEU. This is particularly pronounced at 1-shot, where the difference
in both entailment and contradiction remains significant after correction
for multiple comparisons. Together with the manual evaluation, these
results suggest that task adaptation can improve agreement with
reference descriptions without necessarily reducing factual errors.
Reference-based metrics alone are therefore insufficient for evaluating
faithfulness in table-to-text generation.

The qualitative analysis further supports this observation. Rather than
simply differing in the frequency of hallucinations, the evaluated
models also differ in their failure modes. The autoregressive baselines
frequently produce severe failures such as topic drift, prompt leakage,
or unrelated generations. In contrast, LLaDA-based generations
generally remain more closely aligned with the input table and produce
fluent descriptions, while their remaining errors are often localized,
such as entity substitutions. QLoRA does not eliminate these errors
and, in our manual sample, increases the frequency of entity
substitutions. These findings motivate evaluating not only whether a
generation is incorrect, but also the type and severity of its factual
errors.

Our study nevertheless presents limitations. First, our experiments are conducted exclusively on ToTTo; extending this evaluation to other structured generation benchmarks (e.g. WikiBio, domain-specific data-to-text tasks) would test the generality of our findings.

Second, our comparison is limited to a small set of open-source models reproducible on a single GPU; larger and instruction-tuned systems were excluded to preserve reproducibility. Our conclusions are therefore specific to the models studied here and should not be extrapolated to the autoregressive or diffusion paradigms as a whole.

Finally, faithfulness itself remains an open evaluation problem. While NLI-based entailment captures aspects of factual consistency that lexical metrics fail to detect, our qualitative analysis shows that fluent generations containing subtle entity substitutions may still receive relatively high automatic scores. Developing evaluation protocols capable of distinguishing different types of factual errors therefore remains an important direction for future research.

\paragraph{Information coverage.}
Our manual evaluation focuses on factual errors
expressed in generated text rather than exhaustive
coverage of source information. Consequently,
the reported categories do not directly quantify
omissions of relevant table content. Evaluating
information coverage alongside factual consistency
remains an important direction for future work.

\section{Conclusion}

We presented a diagnostic study of autoregressive and diffusion
language models for table-to-text generation on ToTTo. LLaDA-based
models achieve substantially stronger reference-based performance than
the considered GPT-2, GPT-Neo, and MDLM baselines, with
LLaDA+QLoRA obtaining the highest aggregate BLEU score.

However, source-grounded NLI evaluation, statistical testing, and
manual error analysis show that stronger reference similarity does not
systematically imply greater factual faithfulness. In particular, QLoRA
improves aggregate lexical scores without providing statistically
significant gains over pretrained LLaDA on the tested sentence-level
reference metrics, while faithfulness can deteriorate in selected
settings. These results highlight the need to evaluate structured text
generation using source-grounded factuality measures alongside
conventional reference-based metrics.

\begingroup
\hbadness=2000
\bibliography{custom}
\endgroup
\appendix

\section{Main Results}
Table~\ref{tab:all_results} reports the complete quantitative results
across models and ICL settings.
Table~\ref{tab:qualitative} presents a qualitative comparison on a representative ToTTo example.
Hallucinated information is highlighted in \textcolor{red}{red}, while repetitive or degenerate generations are highlighted in \textcolor{blue}{blue}.

Table~\ref{tab:failure_analysis} highlights the failure analysis on a
challenging ToTTo example. Although all models identify the movie \textit{Kids in Love}, they hallucinate the actress playing the role of Evelyn or drift away from the table content.

Table~\ref{tab:human_eval} presents results of the uman evaluation over 30 examples per model at $k=3$.

\begin{table*}[t]
\centering
\caption{Quantitative results across all models and in-context learning settings on ToTTo. Higher values are better for BLEU, ROUGE-L, BERTScore, and NLI entailment, while lower Self-BLEU indicates greater output diversity.}
\label{tab:all_results}
\small
\setlength{\tabcolsep}{5pt}
\begin{tabular}{llccccc}
\toprule
\textbf{Model} & \textbf{Shot} &
\textbf{BLEU $\uparrow$} &
\textbf{R-L $\uparrow$} &
\textbf{BERTScore $\uparrow$} &
\textbf{NLI Ent. $\uparrow$} &
\textbf{Self-BLEU $\downarrow$} \\
\midrule

\multirow{5}{*}{GPT-2}
& 0  & 3.45 & 0.130 & 0.821 & 0.518 & 0.188 \\
& 1  & 4.64 & 0.212 & 0.837 & 0.239 & 0.085 \\
& 3  & 2.92 & 0.180 & 0.826 & 0.297 & 0.134 \\
& 5  & 2.90 & 0.181 & 0.828 & 0.265 & 0.105 \\
& 10 & 2.40 & 0.171 & 0.825 & 0.302 & 0.129 \\
\midrule

\multirow{5}{*}{GPT-Neo}
& 0  & 2.61 & 0.147 & 0.833 & 0.443 & 0.113 \\
& 1  & 5.74 & 0.201 & 0.834 & 0.317 & 0.092 \\
& 3  & 3.99 & 0.208 & 0.834 & 0.309 & 0.125 \\
& 5  & 3.21 & 0.182 & 0.829 & 0.242 & 0.133 \\
& 10 & 3.54 & 0.192 & 0.834 & 0.172 & 0.170 \\
\midrule

\multirow{5}{*}{MDLM}
& 0  & 4.73 & 0.207 & 0.863 & 0.274 & 0.206 \\
& 1  & 5.22 & 0.221 & 0.865 & 0.193 & 0.132 \\
& 3  & 6.79 & 0.234 & 0.870 & 0.220 & 0.110 \\
& 5  & 6.70 & 0.241 & 0.871 & 0.215 & 0.104 \\
& 10 & 7.93 & 0.267 & 0.875 & 0.191 & 0.088 \\
\midrule

\multirow{5}{*}{LLaDA}
& 0  & 11.25 & 0.355 & 0.895 & 0.781 & 0.121 \\
& 1  & 21.54 & 0.459 & 0.913 & 0.623 & 0.079 \\
& 3  & 21.83 & 0.468 & 0.918 & 0.615 & 0.079 \\
& 5  & 21.96 & 0.468 & 0.917 & 0.516 & 0.088 \\
& 10 & 20.73 & 0.470 & 0.918 & 0.565 & 0.103 \\
\midrule

\multirow{5}{*}{LLaDA + QLoRA}
& 0  & 12.09 & 0.374 & 0.900 & 0.726 & 0.109 \\
& 1  & 24.15 & 0.474 & 0.914 & 0.355 & 0.059 \\
& 3  & 23.28 & 0.474 & 0.916 & 0.531 & 0.061 \\
& 5  & 23.91 & 0.472 & 0.916 & 0.527 & 0.077 \\
& 10 & 21.99 & 0.477 & 0.919 & 0.552 & 0.073 \\

\bottomrule
\end{tabular}
\end{table*}

\begin{table*}[t]
\centering
\scriptsize
\caption{
Qualitative comparison on a representative ToTTo example.
Hallucinated information is highlighted in \textcolor{red}{red}, while repetitive or degenerate generations are highlighted in \textcolor{blue}{blue}.
}
\label{tab:qualitative}

\renewcommand{\arraystretch}{1.18}

\begin{tabular}{p{0.17\linewidth}p{0.77\linewidth}}
\toprule

\textbf{Input Table}
&
Argentine films of 1950 : Don Fulgencio
\\

\midrule

\textbf{Reference}
&
``Don Fulgencio'' is a 1950 Argentinian film.
\\

\midrule

\textbf{GPT-2 (1-shot)}
&
Don Fulgencio,
\textcolor{red}{a former professional football player from Argentina,
was a member of the Argentine Football Federation (AFF) and played for the
Argentine Football Federation (AFF) from 1950 to 1960.}
\textcolor{red}{Table: Argentine films of 1960 : Don Fulgencio Description: Don Fulgenc}
\\

\midrule

\textbf{GPT-Neo (3-shot)}
&
Don Fulgencio is
\textcolor{red}{a film directed by Alberto Lecchi.}
\textcolor{red}{Table: col\_1 : 0.1 (32.2) | col\_7 : 0.1 (32.2) | col\_13 : 0.1 (32.2)}
\textcolor{red}{Description: The film is a}
\\

\midrule

\textbf{MDLM (10-shot)}
&
Don Fulgencio is a 1950 Argentine film
\textcolor{red}{directed by Luis Bayón},
\textcolor{blue}{directed by Luis Bayón, who by the, directed, the,, Don,
Fulgencio, the, Don, the,, Don, the, the, the, the, Don, the, the, the,
the, the, the, the, the, the, the, the, the, the, the, the, the, the,
the, the.}
\\

\midrule

\textbf{LLaDA (5-shot)}
&
Don Fulgencio is a 1950 Argentine film.
\\

\midrule

\textbf{LLaDA + QLoRA (1-shot)}
&
Don Fulgencio is a 1950 Argentine film.
\\

\bottomrule

\end{tabular}
\end{table*}

\begin{table*}[t]
\centering
\scriptsize
\caption{Failure analysis on a challenging ToTTo example. Although all models identify the movie \textit{Kids in Love}, they hallucinate the actress playing the role of Evelyn or drift away from the table content.}
\label{tab:failure_analysis}

\begin{tabular}{p{2.2cm}p{12.8cm}}
\toprule
\textbf{Input Table} &
Year: 2016 \;|\; Title: Kids in Love \;|\; Role: Evelyn \\

\midrule

\textbf{Reference} &
Alma Jodorowsky had the role of Evelyn in the 2016 film \textit{Kids in Love}. \\

\midrule

GPT-2 &
\bad{Generates an unrelated description about the population of Hardyston township}, completely ignoring the input table. \\

\addlinespace

GPT-Neo &
Correctly recognizes the movie title but \bad{continues by generating another unrelated table and historical information}, resulting in severe prompt leakage and hallucination. \\

\addlinespace

MDLM &
Produces
\bad{``Olivia Wilson played the role of Evelyn''}
and additionally hallucinates
\bad{a fictional film ``The Children in Love''}. \\

\addlinespace

LLaDA &
Produces
\bad{``Maisie Williams starred as Evelyn''}.
The sentence is fluent but replaces the correct actress with a hallucinated person. \\

\addlinespace

LLaDA + QLoRA &
Produces
\bad{``Colin Hanks starred as Evelyn''}.
The output is concise and grammatical, but still hallucinates the actress despite correctly preserving the movie title and year. \\

\bottomrule
\end{tabular}

\end{table*}

\begin{table}[h]
\centering
\caption{Human evaluation results over 30 examples per model at $k=3$ 
         (percentages of outputs exhibiting each error type).}
\label{tab:human_eval}
\resizebox{\columnwidth}{!}{%
\begin{tabular}{lcccc}
\toprule
\textbf{Model} & \textbf{Faithful (\%)} & \textbf{Entity Subst. (\%)} 
               & \textbf{Topic Drift (\%)} & \textbf{Unsupported (\%)} \\
\midrule
GPT-2               & 0.00  & 20.00 & 40.00 & 30.00 \\
GPT-Neo   & 0.00  & 14.00 & 30.00 & 20.00 \\
LLaDA               & 47.00 & 23.00 & 10.00 & 40.00 \\
LLaDA + QLoRA       & 26.67 & 40.00 & 10.00 & 37.00 \\
\bottomrule
\end{tabular}%
}
\end{table}

\section{Prompt Template}
\label{appendix:prompt}

All models were evaluated under the same prompting protocol to ensure a fair comparison between autoregressive and diffusion-based language models. The only experimental variable was the underlying language model, while the prompt structure, retrieved demonstrations, and evaluation protocol remained identical.

The system instruction used throughout all experiments is reported below.

\begin{quote}
\small
\texttt{
You are an expert at generating factual natural language descriptions from structured tables.
}
\end{quote}

For the zero-shot setting, the model receives only the system instruction followed by the serialized input table.

\begin{quote}
\small
\begin{verbatim}
You are an expert at generating factual natural
language descriptions from structured tables.

Generate one faithful description for the following table.

Table:
<serialized_table>

Description:
\end{verbatim}
\end{quote}

For the few-shot settings ($k \in \{1,3,5,10\}$), demonstrations are inserted before the target table. Each demonstration consists of a serialized table paired with its corresponding human-written reference description.

\begin{quote}
\small
\begin{verbatim}
You are an expert at generating factual natural
language descriptions from structured tables.

Examples:

Example 1
Table:
<serialized_table_1>

Description:
<reference_1>

...

Example k
Table:
<serialized_table_k>

Description:
<reference_k>

Generate one faithful description 
for the following table.

Table:
<target_serialized_table>

Description:
\end{verbatim}
\end{quote}

Retrieved demonstrations are selected using dense retrieval. Each serialized table is embedded using the \texttt{sentence-transformers/all-MiniLM-L6-v2} encoder. Cosine similarity is then computed between the target table and all training examples, and the top-$k$ nearest neighbours are selected. To ensure fair comparisons across in-context learning settings, nested retrieval is employed: the 3-shot prompt contains the first three examples of the 10-shot retrieval set, while the 5-shot prompt extends the same ordering. Consequently, all models receive exactly the same demonstrations for a given evaluation example.

Finally, all prompts terminate with the keyword \texttt{Description:}, leaving the model responsible for generating a single factual natural language description without additional explanations or formatting.

\section{Experimental Setup \& Hyperparameters}
\label{appendix:setup}

This appendix reports the implementation details and hyperparameters used throughout all experiments. Unless otherwise specified, identical experimental settings were adopted for all models to ensure a fair comparison.

\subsection{Dataset}

All experiments were conducted on the ToTTo benchmark. The evaluation protocol follows the official train/evaluation split. During inference, only the evaluation subset was used for testing, while the training subset served exclusively as the retrieval corpus for in-context learning.

For each evaluation example, dense retrieval was performed over the training set to construct nested in-context demonstrations for the 1-shot, 3-shot, 5-shot and 10-shot settings.

\subsection{Dense Retrieval}

Relevant demonstrations were retrieved using the pretrained encoder
\texttt{sentence-transformers/all-MiniLM-L6-v2}. Each serialized table was encoded into a dense embedding, L2-normalized, and cosine similarity was computed against all training examples. The top-$k$ nearest neighbours were selected to construct the in-context examples.

Nested retrieval was employed so that larger prompt configurations always contained the demonstrations used by smaller configurations.

\subsection{Inference Hyperparameters}

Table~\ref{tab:appendix_inference} summarizes the generation parameters used for LLaDA inference.

\begin{table}[h]
\centering
\caption{Inference hyperparameters used for pretrained LLaDA.}
\label{tab:appendix_inference}
\small
\begin{tabular}{lc}
\toprule
Parameter & Value \\
\midrule
Model & LLaDA-8B-Instruct \\
Random seed & 42 \\
Generation length & 64 tokens \\
Diffusion steps & 64 \\
Block length & 32 \\
Temperature & 0.0 \\
Classifier-free guidance & 0.0 \\
Remasking strategy & Low-confidence \\
Retrieval encoder & all-MiniLM-L6-v2 \\
Prompt settings & 0 / 1 / 3 / 5 / 10-shot \\
\bottomrule
\end{tabular}
\end{table}

All experiments were executed with a fixed random seed in order to ensure deterministic retrieval and generation whenever possible.

\subsection{QLoRA Fine-Tuning}

Parameter-efficient fine-tuning was performed using QLoRA. Only low-rank adaptation matrices were optimized while the pretrained model weights remained frozen. This considerably reduced the memory requirements while preserving the ability to adapt the model to the ToTTo task.

Table~\ref{tab:appendix_qlora} reports the training hyperparameters.

\begin{table}[h]
\centering
\caption{QLoRA fine-tuning hyperparameters.}
\label{tab:appendix_qlora}
\small
\begin{tabular}{lc}
\toprule
Parameter & Value \\
\midrule
LoRA rank ($r$) & 16 \\
LoRA alpha & 32 \\
LoRA dropout & 0.05 \\
Learning rate & $2\times10^{-4}$ \\
Optimizer & AdamW \\
Scheduler & Linear \\
Warmup ratio & 0.03 \\
Epochs & 1 \\
Per-device batch size & 1 \\
Gradient accumulation & 16 \\
Weight decay & 0.01 \\
Maximum sequence length & 1024 \\
Precision & bfloat16 \\
Quantization & 4-bit NF4 \\
\bottomrule
\end{tabular}
\end{table}

\subsection{Hardware}

All experiments were conducted on a single NVIDIA RTX 4050 Laptop GPU. Diffusion inference for LLaDA was performed using 8-bit quantization together with automatic CPU offloading in order to fit the 8B model within the available GPU memory. The retrieval encoder and evaluation metrics were executed on the same hardware.

The source code fixes the random seed to 42 and uses identical retrieval examples across all compared models, ensuring that the language model itself remains the only experimental variable.
\section{Evaluation Metrics}
\label{appendix:metrics}

Generation quality is evaluated using complementary metrics measuring lexical overlap, semantic similarity, diversity and factual consistency. All metrics are computed on the evaluation set using identical prediction files for every model.

\subsection{BLEU}

We report SacreBLEU, which computes modified $n$-gram precision together with a brevity penalty. Higher BLEU indicates greater lexical agreement between the generated description and the reference.

\subsection{ROUGE}

We report ROUGE-L, which measures the longest common subsequence (LCS) shared by the generated and reference descriptions. Higher values indicate better structural agreement.

\subsection{BERTScore}

BERTScore evaluates semantic similarity using contextual embeddings from a pretrained Transformer encoder. We report the F1 score averaged over the evaluation set. Higher values indicate stronger semantic similarity.

\subsection{Self-BLEU}

Self-BLEU measures generation diversity by treating each generated description as a hypothesis and all remaining generations as references.

\[
\mathrm{SelfBLEU}
=
\frac{1}{N}
\sum_{i=1}^{N}
\mathrm{BLEU}
\left(
g_i,
\{g_j\}_{j\neq i}
\right).
\]

Lower Self-BLEU indicates greater diversity, whereas higher values suggest repetitive outputs.

\subsection{Natural Language Inference}

To assess source faithfulness, we employ a Natural Language Inference
(NLI) model. The serialized source table is treated as the premise
and the generated description as the hypothesis. We report the mean
entailment and contradiction probabilities over the evaluation set.

Higher entailment indicates that the generated content is better
supported by the source table, whereas higher contradiction indicates
stronger factual inconsistency with the structured input. Unlike
reference-based metrics, this formulation directly evaluates whether
the information expressed by the generation is supported by its
source.
\subsection{Complementary Evaluation}

Each metric captures a different aspect of generation quality. BLEU and ROUGE evaluate lexical overlap, BERTScore measures semantic similarity, Self-BLEU quantifies output diversity, and NLI estimates factual consistency. Their combination provides a more complete assessment than any individual metric alone and supports the analysis presented in Section~3.
\section{Additional Qualitative Examples}
\label{app:additional_examples}

We provide two additional examples illustrating characteristic
generation behaviors observed during manual analysis. The first
shows a case in which LLaDA-based models remain closely grounded
in the source table, while the second illustrates a more subtle
failure mode in which the overall structure is preserved but a
specific entity is hallucinated.

\subsection{Grounded Generation}

\noindent\textbf{Input table:}
\begin{quote}
Argentine films of 1950 : Don Fulgencio
\end{quote}

\noindent\textbf{Reference:}
\begin{quote}
Don Fulgencio is a 1950 Argentinian film.
\end{quote}

\noindent\textbf{LLaDA:}
\begin{quote}
Don Fulgencio is a 1950 Argentine film.
\end{quote}

\noindent\textbf{LLaDA + QLoRA:}
\begin{quote}
Don Fulgencio is a 1950 Argentine film.
\end{quote}

Both LLaDA-based models produce a concise description that preserves
the information contained in the source table. The minor lexical
variation between ``Argentinian'' in the reference and ``Argentine''
in the generations does not alter the underlying fact. This example
illustrates a case in which reference similarity and source
faithfulness are aligned.

\subsection{Localized Entity Hallucination}

\noindent\textbf{Input table:}
\begin{quote}
Year : 1987 \textbar{} Title : Irupatham Noottandu
\textbar{} Role : Sagar Alias Jacky
\end{quote}

\noindent\textbf{Reference:}
\begin{quote}
Mohanlal portrayed Sagar Alias Jacky in Irupatham Noottandu (1987).
\end{quote}

\noindent\textbf{LLaDA (5-shot):}
\begin{quote}
In 1987, \textcolor{red}{Sivaji Ganesan} played the role of
Sagar Alias Jacky in the film Irupatham Noottandu.
\end{quote}

\noindent\textbf{LLaDA + QLoRA (5-shot):}
\begin{quote}
In 1987, \textcolor{red}{Suresh Rajikumar} played the role of
Sagar Alias Jacky in the film Irupatham Noottandu.
\end{quote}

Both generations preserve the year, film title, and role, but
introduce an unsupported actor identity. LLaDA substitutes
``Sivaji Ganesan'' for the reference actor Mohanlal, while
LLaDA+QLoRA generates ``Suresh Rajikumar.'' The source table itself
does not provide an actor name, making both additions unsupported.
The source-grounded NLI evaluation accordingly assigns contradiction
probabilities above 0.999 to both generations. This case illustrates
a characteristic localized factual error: the generation remains
topically and structurally aligned with the table while introducing
a specific unsupported entity.

\end{document}